\documentclass{article}
\usepackage{arxiv}
\pdfoutput=1
\usepackage[utf8]{inputenc} 
\usepackage[T1]{fontenc}    
\usepackage{hyperref}       
\usepackage{url}            
\usepackage{booktabs}       
\usepackage{amsfonts}       
\usepackage{nicefrac}       
\usepackage{microtype}      
\usepackage{lipsum}
\usepackage{graphicx}
\graphicspath{{./images/}}
\usepackage{amsmath}

\title{Towards a New Understanding of the Training of Neural Networks with Mislabeled Training Data }

\author{Herbert Gish\\
 Raytheon BBN Technologies\\
  10 Moulton Street, Cambridge, MA 02138\\
  \AND
   Jan Silovsky \thanks{Now at Apple, Cambridge, MA} \\
  Raytheon BBN Technologies  \\
10 Moulton Street, Cambridge, MA 02138 \\
 \And
 Man-Ling Sung\\
 Chinese University of Hong Kong\\
 Hong Kong, China \\
\And
 Man-Hung Siu \footnotemark[1]\\
 Raytheon BBN Technologies\\
 10 Moulton Street, Cambridge, MA 02138\\
\And
 William Hartmann \\
 Raytheon BBN Technologies\\
 10 Moulton Street, Cambridge, MA 02138\\
\And
 Zhuolin Jiang\\
 Raytheon BBN Technologies\\
 10 Moulton Street, Cambridge, MA 02138\\
}

\begin{document}
\maketitle
\begin{abstract}
We investigate the problem of machine learning with mislabeled training data.  We try to make the effects of mislabeled training better understood through analysis of the basic model and equations that characterize the problem.  This includes results about the ability of the noisy model to make the same decisions as the clean model and the effects of noise on model performance.  In addition to providing better insights we also are able to show that the Maximum Likelihood (ML) estimate of the parameters of the noisy model determine those of the clean model.  This property is obtained through the use of the ML invariance property and leads to an approach to developing a classifier when training has been mislabeled: namely train the classifier on noisy data and adjust the decision threshold based on the noise levels and/or class priors.   We show how our approach to mislabeled training works with multi-layered perceptrons (MLPs).

\end{abstract}


\large
\section{Introduction}
The problem of mislabeled data occurs when we have a set of data, $(y_i,x_i), i=1,...n$ with $x_i$ the features and labels $y_i$  and some of the labels have been assigned incorrectly.  The mislabeling will, of course, affect the training of a classifier based on observed features.  Our goal is to better understand the effects of mislabeling and how to train classifiers with mislabeled data (which we refer to as noisy data). We assume that the label errors occur randomly, independent of the features, and that the mislabeling rates may be label dependent.

We start with a view of the problem based on the relationship between the posterior probabilities of the noisy labels and the posterior probabilities of the clean labels. We assume we have perfect knowledge of both models. We refer to this relationship as the basic model or equation as given by Equations \ref{eq:1} and \ref{eq:2}.  From the basic model we show how the noisy model can make the same decisions as the clean model and will also show how the noise can make training less efficient directly from the linear relationship between the noisy and clean models.

While the basic model can provide insights into the general properties of the effect of mislabeled training we, of course, need to consider the situation of finite training.  Through the
use of the Maximum Likelihood invariance property and the basic equation, we show that the Maximum Likelihood estimate (MLE) of the clean model is a function of the MLE of the noisy model trained on noisy data, with the important difference being that when a class decision is made by the noisy model it uses a decision threshold determined by the noise parameters.  In essence the MLE models follow the relationship of the basic model that relates the posterior probabilities of the noisy and clean data.   Our analysis shows that the ML estimates of the models do not depend on the noise parameters and if there is a need for their estimation it can be dealt with after the models have been trained. Concisely, this means that we can train our model on the noisy data and then consider the setting of the decision threshold which is determined by the noise parameters and prior probabilities. Alternatively the noisy model can be transformed into the clean model and decisions made with a clean model using a decision threshold of $1/2$, an approach that we consider briefly in  this paper.

 The basic model relates the models for the noisy and the clean data, however the noisy model that is specified by the basic equation is, by definition, meant to model noisy observations, and it is based on the probabilities generated by the clean model followed by noisy transitions. However, the basic equation does not deal with training a specific noisy model from noisy training data.  \it So typically, we will be training a model with class priors that depend on the noisy training data as well as the priors of the clean data. \rm That is, due to noise flipping the labels, the training data may not have the same class priors as the data that has not been mislabeled, which can lead to degraded performance if ignored. We show how to compensate the MLP for the incorrect prior by adjusting the decision threshold employed with the model. We illustrate, via simulated classification experiments, several results from our work.  In particular, in Section 5, the relationship between the amount of training and performance, and the effect of prior mismatch on classification performance.  All experiments were performed with an MLP classifier.

\subsection{Related Work}
There is a large literature on the noisy label problem (see [\ref{Frenay}]) with many papers dealing with issues of classifier robustness and data cleansing procedures and the effect of noisy labels on classifiers other than neural networks.  Of this large literature on the noisy label problem there are a few papers that are closely related to our work.
In [\ref{Natarajan}] the relationship between the clean and noisy model is described but we believe our approach is more intuitive and it also provides us with insight into performance and how it relates to noise levels and amount of training.  Also in [\ref{Natarajan}] they approach the noisy data problem as a weighted error and emphasize SVM training rather than neural networks.  They deal with the issue of unknown noise parameters as well as any impact unknown priors might have by "tuning" a decision threshold via cross-validation and does not explicitly relate the noise and prior parameters to the decision threshold for the classifier.

Some recent papers [\ref{Bekker},\ref{Goldberger},\ref{Sukhbaatar}] have employed training methods for the noisy label problem that have tried to jointly optimize the parameters of the classifier (neural network) along with the noise parameters. These works recognize that their training procedures require good model initializations to be successful. In our view of the problem, as we have already noted, the noise parameters are not part of the Maximum Likelihood optimization process and should be treated separately.  In [\ref{Mnih}, \ref{Bekker}, \ref{Goldberger}, \ref{Sukhbaatar}] all try to estimate neural network models of the clean probability using the cross entropy criterion and none consider the noisy model approach. In the paper [\ref{Menon}] the goal is to find training criteria that are insensitive to the noise corruption parameters.  It also recognizes that class priors do change with noise though it does not examine its effect on neural network training.

\subsection{Our Contributions}
While the basic model is well known (e.g., it is important in  [\ref{Mnih}, \ref{Goldberger},\ref{Sukhbaatar}], as well as others), we have viewed it in new ways to improve our understanding of how:
\begin{itemize}
    \item the noisy model can make the same decisions as the clean model
    \item noise can decrease the efficiency of estimating the model parameters
    \item sufficient noisy training can overcome the effects of the noise
\end{itemize}

In addition we have shown:
\begin{itemize}
    \item how to account for prior shift in the training of MLP's
    \item that prior shifts between between training and test data will occur and can lead to performance degradation if not taken into account during training.
\end{itemize}
and also what maybe most important,
\begin{itemize}
    \item ML or minimum cross entropy training can be accomplished by training a model on noisy data and then correcting for biases introduced by the noise.
\end{itemize}

\section{A model with perfect knowledge}
\label{Perfect}
Our data consists of class labels and features of the form $(y_i,x_i), i=1,\ldots,n.$ where $x_i$ is a feature vector and $y_i$ takes on the values of 1 or 0, denoting which of two classes $x_i$ is associated with.  The heart of the mislabeling problem is that for each $x_i$ we do not actually observe $y_i$ but $z_i$, which is a randomly flipped version of $y_i$.  We assume that this flipping is not dependent on the $x_i$ and the flipping is independent from observation to observation.  In the following we will first consider the case where we have complete knowledge of the pdf's for each of the classes as well as the probabilities of flipping.  From this case we can understand the consequences of mislabeling when model estimation is not of concern.

If we let $z_i$ denote the observed, and possibly mislabeled, class label associated with feature vector $x_i$, we can write the posterior probabilities,
\begin{equation} \label{eq:1}
    \tilde{p}(z_i=1|x_i)=(1-\gamma_1)p(y_i=1|x_i)+\gamma_0p(y_i=0|x_i)
    \end{equation}
where
$\gamma_1$ is the probability that a class 1 label is flipped to a class 0 label and $\gamma_0$ is the probability that a class 0 label is flipped to a 1.
In a similar manner we have,
\begin{equation} \label{eq:2}
\tilde{p}(z_i=0|x_i)=\gamma_1p(y_i=1|x_i)+(1-\gamma_0)p(y_i=0|x_i)
\end{equation}
We will refer the the above equations \ref{eq:1} and \ref{eq:2} as the basic equations of the basic model for mislabeled binary training.

We note that,

\begin{equation} \label{eq:3}
    p(y_i=1|x_i)=1-p(y_i=0|x_i)
\end{equation}
and that
\begin{equation} \label{eq:4}
  \tilde{p}(z_i=1|x_i)=1-\tilde{p}(z_i=0|x_i)  
\end{equation}

Simple substitutions applied to the above equations gives us,
\begin{equation} \label{eq:5}
    \tilde{p}(z_i=1|x_i)=(1-\gamma_1-\gamma_0)p(y_i=1|x_i)+\gamma_0.
\end{equation}

We see that $\tilde{p}$  will be a monotonically increasing function of $p$ provided $\gamma_1+\gamma_0<1$.
This shows us that the posterior probability for  making a decision about the noisy labeled data is linearly related to the posterior probability for making decisions about the noise free data.  Assuming we have the model for the noisy data, i.e., $\tilde{p}(z_i=1|x_i)$ and we wish to make decisions about clean data all we need is to use the correct threshold for $\tilde{p}(z_i=1|x_i)$ which will be the threshold that corresponds to $p(y_i=1|x_i)$ being equal to $1/2$.  That is, since the threshold of $1/2$ is the best (Bayes minimum error) threshold we would use with the clean model we can equivalently achieve the same result by using an equivalent threshold for the noisy model, i.e., we can re-write the above equation as:
\begin{equation}
        p(y_i=1|x_i)=\frac{\tilde{p}(z_i=1|x_i)-\gamma_0}{1-\gamma_1-\gamma_0}
        \label{basic clean model}
\end{equation}
Again some simple algebraic manipulation we find the threshold to use with the noisy model to make decisions about noise free data is given by
\begin{equation}
    thr=\frac{(1-\gamma_1+\gamma_0)}{2}
    \label{threshold1}
\end{equation}
We immediately see that if $\gamma_1=\gamma_0$ the threshold of $1/2$ would be employed as if no noise was present.

From the above we see that the noisy model can make the same decisions about noise free data as can the noise free model. At the risk of overstating the obvious the above equations show us that even in the case of severe noise, e.g., $\gamma_1=\gamma_0=0.49$ the noisy model will provide the same decisions as the clean model.  In a certain sense this noisy data problem has an inherent robustness to it in that severe label noise can be overcome if we can train accurate noisy models.  It has been observed [\ref{Rolnick}] that deep learning networks are robust to massive label noise and while this is so, we think---based on our observation that the robustness is inherent in the nature of the label noise problem---that any model or learning procedure that estimates the noisy model with sufficient accuracy would also exhibit robustness. For the noisy label problem modeling the noisy data is especially efficient because it is a step in estimating the clean model. 

However there is no free lunch and we must keep in mind that the above discussion assumes perfect knowledge about the models.  In practice this would only approximately happen when the models used were not mis-specified and there was sufficient data (relative to the separability of the classes) to provide highly accurate estimates.  We can get a better idea of the model behavior, in particular how errors in estimating $\tilde{p}$  effect estimation of the clean posterior $p$, by examination of Equation \ref{basic clean model}.  If $\gamma_1+\gamma_0 < 1$, which we will assume, then the slope of Equation \ref{basic clean model} will be positive. We can then relate the error in $p, \mathrm{ namely} \  \Delta p$, to the estimation error in $\tilde{p},\ \mathrm{namely}\  \Delta \tilde{p}$, via the derivative of $p$ with respect to $\tilde{p}$  as:
\begin{equation}
    \frac{\Delta p}{\Delta\tilde{p}}=\frac{1}{1-\gamma_1-\gamma_0}
\end{equation}

If we define the total noise as  $n=\gamma_1+\gamma_0$ then we see that the error in estimating $p$ depends on the error in estimation of the noisy model and the noise level given by:
\begin{equation}
    \Delta{p}=\frac{\Delta\tilde{p}}{1-n}, \\
    \ \ \ \ \  \mathrm{where} \ 
    \ \ 0<n<1.
\end{equation}

  The noise controls $n$ and large amounts of training data are required to keep $\Delta\tilde{p}$ sufficiently small. Thus we see how the noisy labeling leads to a decrease in learning efficiency by requiring more training to keep $\Delta\tilde{p}$ small in relation to $1-n$. In Figure 1 we have illustrated the relationship between errors in $\tilde{p}$ and how they relate to errors in $p$ and the role that the noise plays. We note that  when $\gamma_1+\gamma_0 = 1$ no learning is possible.  In Section 5 we have used simulations to illustrate the loss of training efficiency is brought about by the label noise.
  
  The above equation also shows that if there is any noise present then the error in $p$ will always be greater than the error in $\tilde{p}$.  In a test situation of making a decision about the label of clean data the noisy model can be used which will have a smaller error than the clean model, however the decision threshold for the noisy model may also have been estimated and can also have an error.  The decision to use the clean model with a known threshold of $1/2$ or the noisy model with a noisy threshold will be problem dependent.  We see that the basic model, while giving us the relationship between the noisy and clean posteriors, needs to be complemented with information regarding model estimation in the finite training sample situation to effectively deal with the mislabeling problem in practical situations.  We will consider this issue in greater detail in Section \ref{Priors}.
  \begin{figure}[h]
      \centering
      \includegraphics[width=12cm]{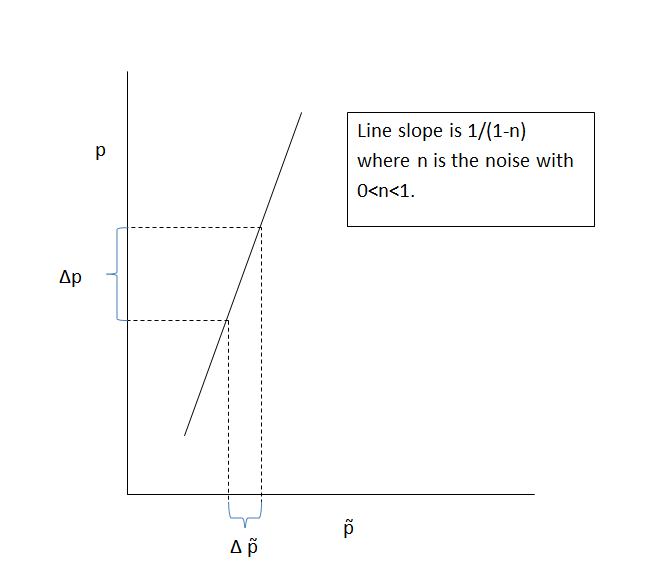}
      \caption{Relationship between errors in $\tilde{p}$, $p$ and the noise n.}
      \label{fig:my_label}
  \end{figure}

\section{Training and Performance Evaluation of Noisy Models}
The models we will consider will be trained with the Maximum Likelihood estimate (MLE) of the model parameters. Namely we have:

\begin{equation}
\mathrm{MLE}= \underset{\theta}{\max}\;\sum_{i=1}^N z_i\mathrm{log}\;(\tilde{p}(z_i=1|x_i,\theta) +(1-z_i)\mathrm{log}(1-\;\tilde{p}(z_i=1|x_i,\theta))
\end{equation}
\label{Invariance}
where \[\tilde{p}(z_i=1|x_i,\theta)\] is our model of the noisy training data. The negative of the MLE is often referred to as the cross-entropy measure.

We denote MLE of the $\theta$ parameters as$\ \hat\theta$ and $\tilde{p}(z_i|\hat\theta,x_i)$ as the MLE of the noisy model.

While the noisy model is fine for prediction of noisy labels our goal is to predict the true label.  For this we need the MLE of $p$ for the clean data.  For this we rely on the Invariance Property of Maximum Likelihood Estimation (MLE) [\ref{Casella and Berger 2002}].  Below we describe how we proceed from the ML estimate of the noisy model to that of the clean model.

\subsection{Posterior probability estimation and the Invariance property of a Maximum Likelihood Estimation (MLE)} The invariance property of MLE states that if $\hat\theta$ is the MLE of $\theta$ then the MLE of the function $g(\theta)$, $\hat g(\theta)$, is equal to $g(\hat\theta)$.  Now consider the estimated posterior model for mislabeled data,\[\tilde{p}(z_i=1|x_i,\hat\theta)\] where $\hat\theta$ are the ML estimated parameters of the neural network for estimation of the probabilities.  We will now consider $\tilde{p}$ as a function of $\hat\theta$, in particular it is a Bernoulli probability  and its Maximum Likelihood estimate is, via the ML invariance property, 
\[\hat{\tilde{p}}(z_i|x_i,\theta)=\tilde{p}(z_i|x_i,\hat\theta),\]
 with $\hat{\tilde{p}}$ being now interpreted as the MLE of the Bernoulli probability $\tilde{p}$ by virtue of the MLE invariance property. 

The reason we wanted to consider $\hat{\tilde{p}}$ as a MLE of a Bernoulli probability is that by the linear relationship given by the noisy and clean probabilities given by the basic equation below 
\begin{equation}
\tilde{p}(z_i=1|x_i,\theta)=(1-\gamma_1)p(y_i=1|x_i,\theta)+\gamma_0(1-p(y_i=1|\theta,x_i))
\end{equation}
we can, with another application of the ML Invariance theorem along with some simple algebra show that the MLE of the clean probability is given by
\begin{equation}
    \hat p =\frac{\hat{\tilde{p}}-\gamma_0}{1-\gamma_1-\gamma_0}
    \label{ML perfect}
\end{equation}
which is the ML version of what we have called the basic model.

There are several inferences we can make from the results of applying the invariance property:
\begin{enumerate}
    \item The ML estimates follow the basic equations
    \item Once the noise model has been trained on the noisy data the only thing left to adjust are decision thresholds. Since we want to classify data whose labels are clean we need to compare $\hat{p}$ to a threshold of $1/2$ which in turn gives us the correct threshold for $\hat{\tilde{p}}$, which is $\frac{1-\gamma_1+\gamma_0}{2}$.  
    \item If the $\gamma$'s are unknown, after MLE is obtained we can treat the $\gamma$'s as variables to be estimated.  We can even change the training criterion from ML to minimum error.  In Section 4, we suggest various approaches to unknown parameter estimation.
\end{enumerate}
\subsection{Insight into the ML Invariance property}
While the invariance property of MLE demonstrates that we can train directly on the noisy data, we can also demonstrate this fact with an alternative approach.  Let us assume we wish to estimate the probability of a Bernoulli random variable, say the probability $p$ of heads in a coin tossing experiment.  We also assume that there has been some mislabeling of the coin tosses in the training data.  We first train a model by the ML criterion to obtain the MLE of $\tilde{p}$, the corrupt model that pays no attention to mislabels. We find this estimate by finding the solutions to,

\begin{equation}
\frac{\mathrm{d log}L(\tilde{p})}{\mathrm{d}\tilde{p}}=0,
\end{equation}
where $\mathrm{log}L(\tilde{p})$ is the log likelhood of the Bernoulli problem. From the basic equation we know that $\tilde{p}$ takes the form,\[\tilde{p}=(1-\gamma_1)p+\gamma_0(1-p).\]
We now have to solve the ML estimation problem for the probability $p$ and so we want to solve,\[\frac{\mathrm{d} log L(\tilde{p}(p))}{\mathrm{d}p}=0,\] and by the chain rule we have,
\begin{equation}\frac{\mathrm{d log }L(\tilde{p}(p))}{ \mathrm{d}p}=\frac{\mathrm{d log} L(\tilde{p}(p))}{\mathrm{d}\tilde{p}}\frac{\mathrm{d}\tilde{p}}{\mathrm{d}p}=0.
\end{equation}

Since $\frac{d\tilde{p}}{dp}$ is a constant  we find that we first need to find the ML solution for $\tilde{p}$ and then use it to solve for $p$.  This gradient based approach gives us the same answer that we got from the invariance property.  Both approaches say first obtain the MLE of the noisy model and then fix it after training.  

There have been some papers recently [\ref{Goldberger},\ref{Sukhbaatar}] that have employed training methods that have tried to deal with the noisy label problem by jointly optimizing the parameters of the classifier (neural network) along with the linear transformation parameters, what we have called the $\gamma$'s.  Based on our understanding of the problem, and from what we saw in Section 3, the transformation parameters are not part of the Maximum Likelihood optimization process and should treated separately.

\section{Training a neural network on mislabeled training data}
\subsection{Introduction}Our approach to training the neural network to deal with the mislabeled training is to follow the results of Section \ref{Invariance}  and initially simply ignore the mislabeling.  After training the noisy model we can use it to make decisions by comparing the posterior probability it has produced to a threshold based on prior probabilities and noise parameters. This threshold adjustment compensates for the biases introduced into to our model by ignoring the noise in the data. More specifically the effect of the noise is explicitly dealt with by its impact on the prior probabilities of the training data.


We also note that in training a neural network classifier we cannot go to the basic model/equations for guidance. The basic equations enable us to establish the relationship between the noisy and clean posterior probabilities and also show us how to make the same decisions with the noisy and clean models but they don't deal with the training of specific models such as a neural network.  In particular the basic equations do not deal with the issue of class priors associated with the clean and noisy data.


 
\subsection{Estimating the model: Threshold selection based on priors and noise parameters}
Whenever our training corpus is corrupted by mislabeling there is the potential that the data counts for each of the classes change in a significant way. This will be especially true when $\gamma_1$ and $\gamma_0$ differ significantly.  When this happens we will be in a situation where the training data has prior class probabilities that differ from data for which class decisions are to be made. If a classifier is trained with the incorrect priors then the prior mismatch can lead to a degradation in performance. The degree to which the mismatch can be compensated will depend on the degree of the mismatch and on the type of classifier being trained.  The prior probabilities that we have for model training depends on the prior for the clean data that we will denote by $P_1$ for class 1 and $P_0=1-P_1$ as the prior for class 0.  We can express the relationship between the priors of the clean data and the noisy training by:

\begin{eqnarray}
\label{Prior Eqn}
    \tilde{P}_1 & = & (1-\gamma_1)P_1 + \gamma_0 P_0, \\
    \tilde{P}_0 &= & \gamma_1P_1+(1-\gamma_0)P_0 \nonumber,
\end{eqnarray}
where $\tilde{P}_1+\tilde{P}_0=1$ and $P_1+P_0=1$.

Since training with the ML criterion requires us to create the noisy model trained with our noisy data it will be a model trained with the above class priors.  However, the data we want to classify will have the clean priors.  Thus the effect of the noise on the training is to treat the model training process as one of prior shift between the training and test domains.  Up to this point our approach is actually model independent.  Below we focus on our approach for MLP training and how readily it accounts for prior shift.

\label{Priors}

The approach we will take with the neural network is based on the results in [\ref{Gish,Siu 1994}]. Let the neural network be an MLP of the form,
\begin{equation}
    p(y=1|x)=\frac{\exp{(\beta+z(x,\theta))}}{1+\exp{(\beta+z(x,\theta))}}
\end{equation}
where $\beta+z(x,\theta)$ is the output of the final layer and the input to the sigmoid and where $\beta$ is a constant term.
From [\ref{Gish,Siu 1994}], if the above equation represents the probability of the model trained on the correct prior (i.e. the test prior, the prior that was not changed due to flipping errors), then the model obtained by training with the incorrect prior will be of the form:
\begin{equation}
    p(y=1|x)=\frac{\exp{(\beta^*+z(x,\theta))}}{1+\exp{(\beta^*+z(x,\theta))}}
\end{equation}
where 
\begin{equation}
    \beta^*=\beta+\Delta
\end{equation}

That is, the only change to the neural network by training with the incorrect prior is a change in the constant term by the amount $\Delta$ and where
\begin{equation}
\label{Delta}
    \Delta=\rm{log}\frac{\tilde{P}_1}{1-\tilde{P}_1}-\rm{log}\frac{P_1}{1-P_1}
\end{equation}
with $\tilde{P}_1$ being the noisy prior used for training and $P_1$ being the priors in the target set to be classified. 

The approach taken to arrive at this result is based on the re-sampling of the training data in a way that the training is using the new priors.  The result shows that for an MLP the change to the model will only be an offset to the neural network final layer output given by $\beta+z(x,\theta)$.  Since where the argument of the exponent of the sigmoid equals zero is the decision boundary for the classifier, the effect of the change in prior is to shift the decision boundary by the amount $\Delta$.

After computing $\Delta$ we have two choices for working with the neural network to classify test data.  We can change the network itself by subtracting $\Delta$ from the exponent of the sigmoid or use the trained network directly with a threshold determined by the priors.  We followed the latter course and 
after some algebra we can show that our MLP model trained on the incorrect prior probability can adjust to the priors of the test data by employing the decision threshold given by:
\begin{equation}
    thr=\frac{\exp{\Delta}}{1+\exp{\Delta}}
    \label{thr}
\end{equation}

and after substituting for $\Delta$ in terms of the priors we arrive at
\begin{equation}
thr= \frac{P_0\tilde{P_1}}{P_1\tilde{P_0}+P_0\tilde{P_1}}
\end{equation}
and for convenience we again show the way the priors are related:
\begin{eqnarray}
     \tilde{P}_1 & = & (1-\gamma_1)P_1 + \gamma_0 P_0, \\
    \tilde{P}_0 &= & \gamma_1P_1+(1-\gamma_0)P_0 \nonumber
\end{eqnarray}

For the case where $P_1=1/2$, i.e., the test data has equal priors and the priors for the training set are determined by the flipping parameters $\gamma_1\  \rm{and}\  \gamma_0$.  This will give us
\begin{eqnarray}
thr & = & \tilde{P_1}\\
    thr & = & \frac{1-\gamma_1+\gamma_0}{2}
\end{eqnarray}
which we see is the prior of the training data.  While this is also the same threshold as given by Equation \ref{threshold1} for the basic equation, we note that for the basic model this threshold is \it{not} \rm dependent on prior probabilities while for our MLP the threshold value holds only when the clean data had equal priors, although we know how to change it for other clean data priors.  

\subsection{Dealing with unknown parameters}
In order to train this neural network classifier we required knowledge of the training prior. We also needed the test prior which we assumed we knew, but in general this may not be known.  The training prior can of course be directly estimated from the training data.  The test prior can be estimated if we know the noise levels and vice versa through Equation \ref{Prior Eqn}.  There are other ways for estimating what we need, keeping in mind that we are ultimately after the choice for the best threshold.  In [\ref{Natarajan}] they discuss tuning the threshold parameter to the problem via cross-validation of on a dataset.  In [\ref{Menon}] it is discussed how the reduction in confidence in labels by the noisy model can lead to estimates of the noise values, $\gamma_1$ and $\gamma_0$.  We also believe that the likelihood can be used for noise level estimation and also that a consistency estimate may be useful for estimating the prior for the clean data.  By consistency we mean that if we train a noisy model with the assumption of a particular value for $P_1$ we would expect that upon decoding test data that an estimated value of the prior should be close to the assumed value. There is also the possibility of Bayesian approaches in which we can postulate distributions for  the unknown parameters.  While interesting and important to the problem, it is beyond the scope of this paper to investigate estimation methods for the prior or noise parameters.

\subsection{Extending the Model}
The model we have discussed above assumes a prior for the clean data and a noisy prior that comes from the clean data and the noise.  It is also assumed that the trained system will be used to classify data that has the clean prior.  We can extend this model and assume the prior for the data that will be classified differs from the clean prior and we can denote this prior by,
\begin{eqnarray}
    \mathrm{Evaluation \ set \ class \ 1\  prior}&=&P_{1,eval} \\
     \mathrm{and, \ \ \ \ \ \ \ \ \ \ \ \ \ \ \ \ \ \ \ \ \ \ \ \ \ \ \ \ \ \ \ } & & \nonumber \\
    \mathrm{evaluation \  set \ class\  0 \ prior}&=&P_{0,eval}
\end{eqnarray}
In this case Equation (\ref{Delta}) becomes
\begin{equation}
        \Delta=\rm{log}\frac{\tilde{P}_1}{1-\tilde{P}_1}-\rm{log}\frac{P_{1,eval}}{1-P_{1,eval}}
\end{equation}
and the threshold for this case is,
\begin{equation}
    thr= \frac{P_{0,eval}\tilde{P_1}}{P_{1,eval}\tilde{P_0}+P_{0,eval}\tilde{P_1}}
    \label{thr2}
\end{equation}

Consider the case when the classification/evaluation set has equal priors, we have

\begin{eqnarray}
    thr &=&\tilde{P_1} \\
    \ &=& (1-\gamma_1)P_1 + \gamma_0 P_0, 
\end{eqnarray}
and we see that for equal evaluation priors the threshold is equal to the prior probability of class 1 of the noisy training data.  

Thus with this extension we can have the training prior, the noisy training prior and the test or evaluation set prior and still readily set the appropriate decision threshold.

Another extension of the model is that it applies to neural networks other than MLPs.  In fact it applies to any feed forward neural network with a sigmoidal output stage.  It is the sigmoid that allows us to deal with prior shift in a rather direct way.

In Section \ref{Experiments} we will perform simulated experiments with mismatched priors and show that failing to correct for the prior shift in training can lead to a significant degradation in classification performance.  We assume noise parameters and priors are known in these experiments.

\section{Experimental Results}
\label{Experiments}
We ran simulations in order to demonstrate some of the important points that we have made.  All the simulations are two class experiments where each class consists of a Gaussian mixture model of two terms.  The Gaussians are two dimensional and each simulation is done with various Gaussian mixtures and performances are averaged over the various experimental runs. In all cases the classifier is an MLP with two hidden layers and 15 nodes in each layer.

In the  first experiment we illustrate the loss of efficiency in training due to noise. The training data had equal class priors and the flipping values of of $\gamma_1$ and $\gamma_0$ were set equal.  These choices for class prior and flipping parameters ensured that the decision threshold for the noisy model would be equal to $1/2$, the same value as for the clean model.  The test data upon which classifier performance was evaluated had equal priors.

In Figure 2 we have plotted the classification performance of classifiers trained with differing amounts of noise and differing amounts of training data. The noise levels for this figure are defined as,
\begin{equation}
    n=\gamma_1+\gamma_0
\end{equation}
\begin{figure}[h]
    \centering
    \includegraphics[width=13cm]{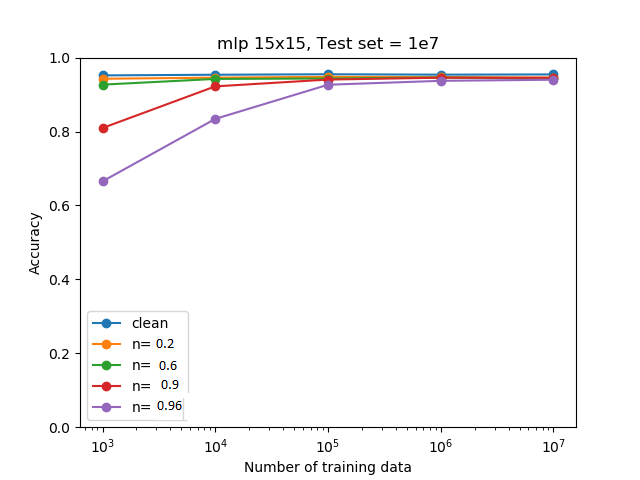}
    \caption{ Classification performance as a function of noise level and amount of training.}
    \label{fig:my_label2}
\end{figure}


We observe, as we have inferred from the basic equations, that more noise can be combated by more training and that with enough training the performance levels approach those of the noise free model.

In Figure 3 we have plotted the classification accuracy  on simulated data again using the two dimensional, two Gaussians per class, mixture models. In the experiments we considered two levels of noise with noise level defined as,
\begin{equation}
    n=\gamma_1+\gamma_0
\end{equation}
as well as various flip ratios defined as $\gamma_0/\gamma_1$. The class priors for the clean data are set to be equal. The solid lines show the performance when the threshold given in Equation \ref{thr} is used and the dotted line showing the performance with a threshold of $1/2$ is employed, ignoring the change in priors due to the noise.  We see that in the low noise case that only a small performance loss occurs when no threshold adjustment is made, and this is for the more extreme flip ratios.  However, in the moderate noise case a significant degradation occurs even at modest flip ratios.  The main takeaway from this experiment is that ignoring the threshold adjustment can lead to significant performance loss.  However, this will depend on the specifics for any given classification situation.
\begin{figure}[h]
    \centering
    \includegraphics[width=15cm]{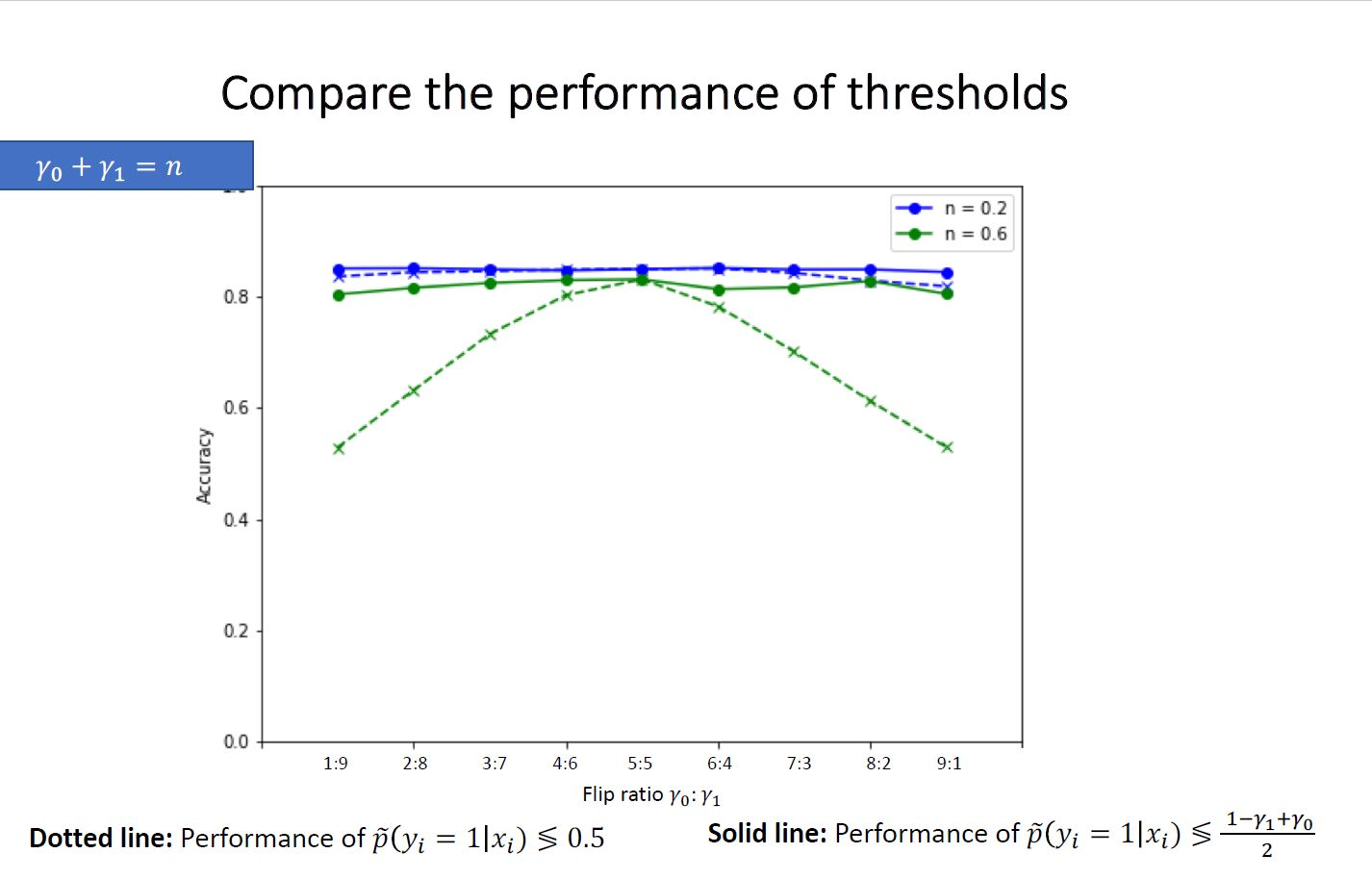}
    \caption{Classifier performance as a function of noise level and flip ratio.}
    \label{fig:my_label3}
\end{figure}


Previous work with neural network based classifiers \cite{Natarajan, Goldberger, Mnih}
have used transformations on estimated posteriors to help combat mislabeling; there doesn't seem to be any threshold adjustment procedure to deal with prior change.  This is an area worthy of closer scrutiny.

\section{Summary and Discussion}
Starting with what we have called the basic model for noisy data we derived the properties that characterize the feature-independent noisy label problem.  These start with the important observation that the noisy model and the clean model will make the same decisions if the noisy model is provided with the appropriate decision threshold.  We also showed how noise makes training less efficient and showed that with a sufficient amount of noisy training that the noisy model would approach the performance of the clean model.  We also derived the ML estimate of the noisy model and showed that the clean ML model  depended on the noisy model just as in the basic equation. The implication is that the estimated noisy model will make the same decision as the estimated clean model provided the noisy model is given the correct threshold.

The basic model, however, does not provide us with guidance in the training of a specific classifier.  For our choice of an MLP neural network classifier and what we have learned about ML estimation, we train the MLP on the noisy data and make adjustments to the decision threshold due to the noise.  We observed that the noise will change the training priors and 
in order to deal with the prior shift we employed the result that MLPs are altered in a very special way by prior shift.  Only a single parameter of the model is changed in a known way by prior shift.  We compensated for the prior shift due to the noise and showed experimental results that ignoring the shift could lead to a degradation in performance.  We also note that our approach to the problem does not make any "corrections" to training data.  The training processes is about estimating the model parameters and dealing with the biases brought about by the mislabeling of data.

Although we have only dealt with the binary problem in this paper we believe the extension to the multi-class case is straight forward.  In particular, we can show that the MLE of the clean model will be the trained noisy model followed by threshold adjustments. Threshold adjustments for mismatched prior probabilities for the multi-class case were given in [\ref{Gish,Siu 1994}].

\bibliographystyle{unsrt}

\end{document}